\documentclass[9pt, conference, compsocconf]{IEEEtran}

\ifCLASSINFOpdf
\else
\fi

\usepackage{times}
\usepackage{subfig}
\usepackage{epsfig}
\usepackage{graphicx}
\usepackage{amssymb}
\usepackage{multirow}
\usepackage{siunitx}
\usepackage{dblfloatfix}
\usepackage{float}
\let\labelindent\relax
\usepackage[inline]{enumitem}
\usepackage{mwe}
\usepackage[export]{adjustbox}
\usepackage{color}
\newcommand{\etal}{\textit{et al. }}
\newcommand{\ie}{\textit{i}.\textit{e. }}
\newcommand{\eg}{\textit{e}.\textit{g. }}
\newcommand{\red}[1]{\textcolor{red}{#1}}
\begin{document}
%
\title{Forensic Scanner Identification Using Machine Learning}


\author{\IEEEauthorblockN{Ruiting Shao and Edward J. Delp}
\IEEEauthorblockA{Video and Image Processing Laboratory (VIPER)\\
School of Electrical and Computer Engineering\\Purdue University\\
West Lafayette, Indiana, USA}
}

\maketitle

\begin{abstract}
Due to the increasing availability and functionality of image editing tools, many forensic techniques such as digital image authentication, source identification and tamper detection are important for forensic image analysis. 
In this paper, we describe a machine learning based system to address the forensic analysis of scanner devices.
The proposed system uses deep-learning to automatically learn the intrinsic features from various scanned images.
Our experimental results show that high accuracy can be achieved for source scanner identification.
The proposed system can also generate a reliability map that indicates the manipulated regions in an scanned image.
\end{abstract}


%
\IEEEpeerreviewmaketitle

\section{Introduction}
With powerful image editing tools such as Photoshop and GIMP being easily accessible, image manipulation has become very easy.
Hence, developing forensic tools to determine the origin or verify the authenticity of a digital image is important. 
These tools provide an indication as to whether an image is modified and the region where the modification has occurred.
A number of methods have been developed for digital image forensics. 
For example, forensic tools have been developed to detect copy-move attacks ~\cite{fridrich2003detection, bayram2009efficient} and splicing attacks ~\cite{shi2007natural}. 
Methods are also able to identify the manipulated region regardless of the manipulation types~\cite{popescu2005exposing, bayar2016deep}.
Other tools are able to identify the digital image capture device used to acquire the image~\cite{lukas2006digital, bayram2005source, tuama2016camera}, which can be a first step in many types of image forensics analysis. 
The capture of ``real'' digital images (not computer-generated images) can be roughly divided into two categories: digital cameras and scanners.

In this paper, we are interested in forensics analysis of images captured by scanners. 
Unlike camera images, scanned images usually contain additional features produced in the pre-scanning stage, such as noise patterns or artifacts generated by the devices producing the ``hard-copy'' image or document. 
These scanner-independent features increase the difficulty in scanner model identification. 
Many scanners also use 1D ``line'' sensors, which are different than the 2D ``area'' sensors used in cameras.
Previous work in scanner classification and scanned image forensics mainly focus on handcrafted feature extraction~\cite{khanna2009scanner, dirik2009flatbed, gloe2007forensics}. 
They extract features unrelated to image content, such as sensor pattern noise~\cite{khanna2009scanner}, dust and scratches~\cite{dirik2009flatbed}. 
In ~\cite{gou2007robust}, Gou \etal extract statistical features from images and use principle component analysis (PCA) and support vector machine (SVM) to do scanner model identification.
The goal is to classify an image based on scanner model rather than the exact instance of the image.
In ~\cite{khanna2009scanner}, linear discriminant analysis (LDA) and SVM are used with the features which describe the noise pattern of a scanned image to identify the scanner model.
This method achieves high classification accuracy and is robust under various post-processing (\eg, contrast stretching and sharpening).
In ~\cite{dirik2009flatbed}, Dirik \etal propose to use the impurities (\ie, dirt) on the scanner pane to identify the scanning device.

Convolutional neural networks (CNNs) such as VGG~\cite{vgg}, ResNet~\cite{resnet}, GoogleNet~\cite{inception}, and Xception~\cite{xception} have produced state-of-art results in object classification on ImageNet~\cite{imagenet_cvpr09}. 
CNNs have large learning capacities to ``describe'' imaging sensor characteristics by capturing low/median/high-level features of images~\cite{tuama2016camera}. 
For this reason, they have been used for camera model identification~\cite{tuama2016camera, bondi2017first} and have achieved state-of-art results.

In this paper, we propose a CNN-based system for scanner model identification.
We will investigate the reduction of the network depth and number of parameters to account for small image patches (\ie, $64 \times 64$ pixels) while keeping the time for training in a reasonable range.
Inspired by ~\cite{xception}, we propose a network that is light-weight and also combines the advantages of ResNet~\cite{resnet} and GoogleNet~\cite{inception}.
The proposed system can achieve a good classification accuracy and generate a reliability map (\ie, a heat map, to indicate the suspected manipulated region).

\section{Proposed System}
\label{sec:system}
\begin{figure}[htb!]
   \begin{minipage}[b]{1.0\linewidth}
      \centering
      \centerline{\includegraphics[width=\linewidth]{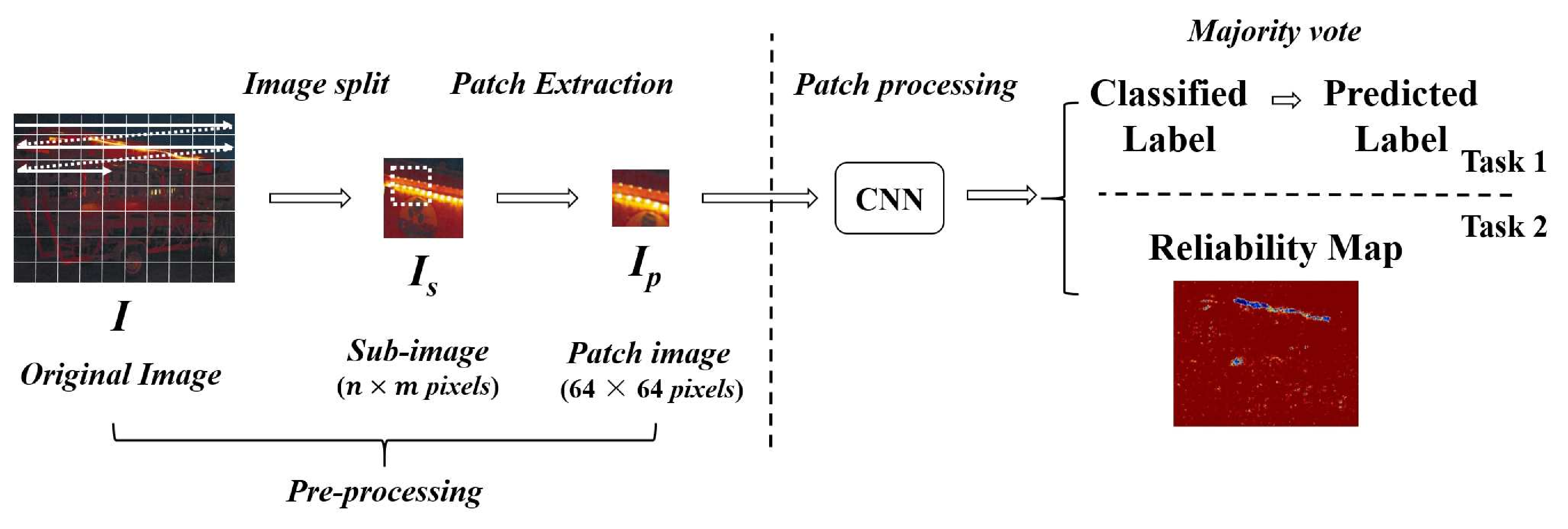}}
   \end{minipage}
   \caption{Our approach for scanner model classification.}\medskip
   \label{fig:system}
\end{figure}

The proposed system is shown in Figure \ref{fig:system}. 
An input image $I$ is first split into smaller sub-images $I_{s}$ of size $n \times m$ pixels. 
This is done for four reasons: 
\begin{enumerate*}[label={\alph*)}]
   \item to deal with large scanned images at native resolution,
   \item to take location independence into account,
   \item to enlarge the dataset, and 
   \item to provide low pre-processing time and memory usage.
\end{enumerate*}

\begin{figure*}[!htb]
    \subfloat[Our Proposed Network \label{fig:patch_incep_res}]{\includegraphics[width=0.7\textwidth,valign=c]{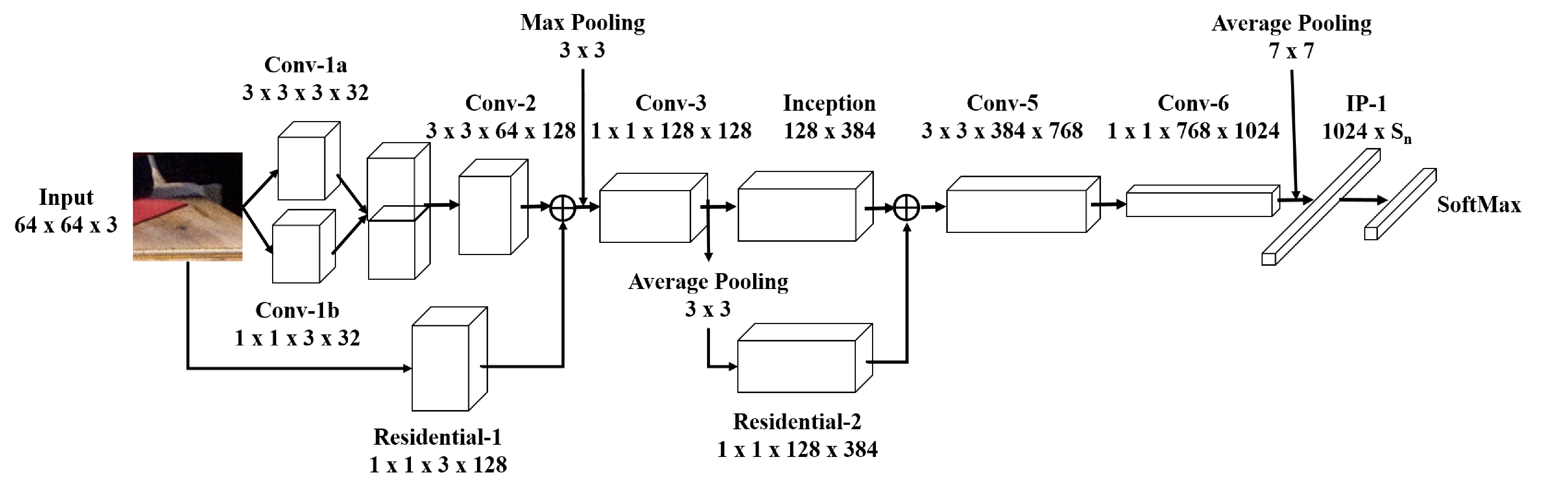}}%
    \quad
    \quad
    \subfloat[The Inception Module \label{fig:incep_module}]{\includegraphics[width=0.2\textwidth,valign=c]{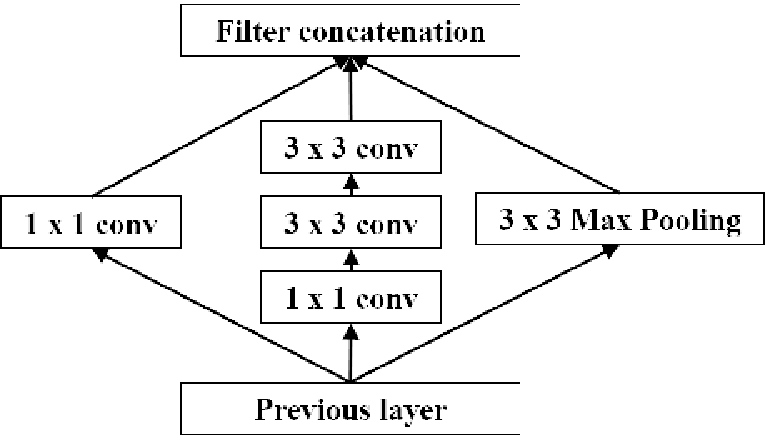}}%

   \caption{(a) is the neural network architecture investigated in this paper. 
     The input image size is $64 \times 64 \time 3$ pixels. 
     The ``IP layer'' is the inner product layer. 
     The residential block is a $1 \times 1$ convolution layer followed by batch normalization. 
     The SoftMax layer acts as a normalized exponential function applied to the layer before the output. 
     The output of SoftMax layer can be interpreted as the probabilities of the input image patch belonging to the scanner models individually.
     The final output layer size is $1 \times N_{s}$, where $N_{s}$ is the number of the scanner models used in the training stage.
     (b) is the inception module used in (a).}\medskip
   \label{fig:networks}
\end{figure*}

\subsection{Training}
\label{ssec:training}
As indicated in Figure \ref{fig:system}, input image $I$ is split into sub-images $I_{s}$ ($n \times m$ pixels) in zig-zag form.
The values of $n$ and $m$ should be no smaller than $64$.
From each $I_{s}$, a patch of size $64 \times 64$ is extracted from a random location.
We denote this extracted patch as $I_{p}$.
These extracted patches $I_{p}$ along with their corresponding labels $S$ are inputs into the network.
This pre-processing enables the proposed system to work with small-size images and use a smaller network architecture to save training time and memory usage.
Designing a suitable network architecture is an important part in the scanner model identification system.
There are several factors that need to be considered to build the network: 
\begin{enumerate*}[label={\alph*)}]
  \item the kernel size,
  \item the utilization of pooling layers,
  \item the depth of the network, and
  \item the implementation of the network modules.
\end{enumerate*}
Our proposed network is shown in Figure \ref{fig:networks}.

\begin{figure}[!hbt]
  \centering

  \subfloat[per patch \label{fig:cm_2_patch}]{\includegraphics[width=0.44\linewidth]{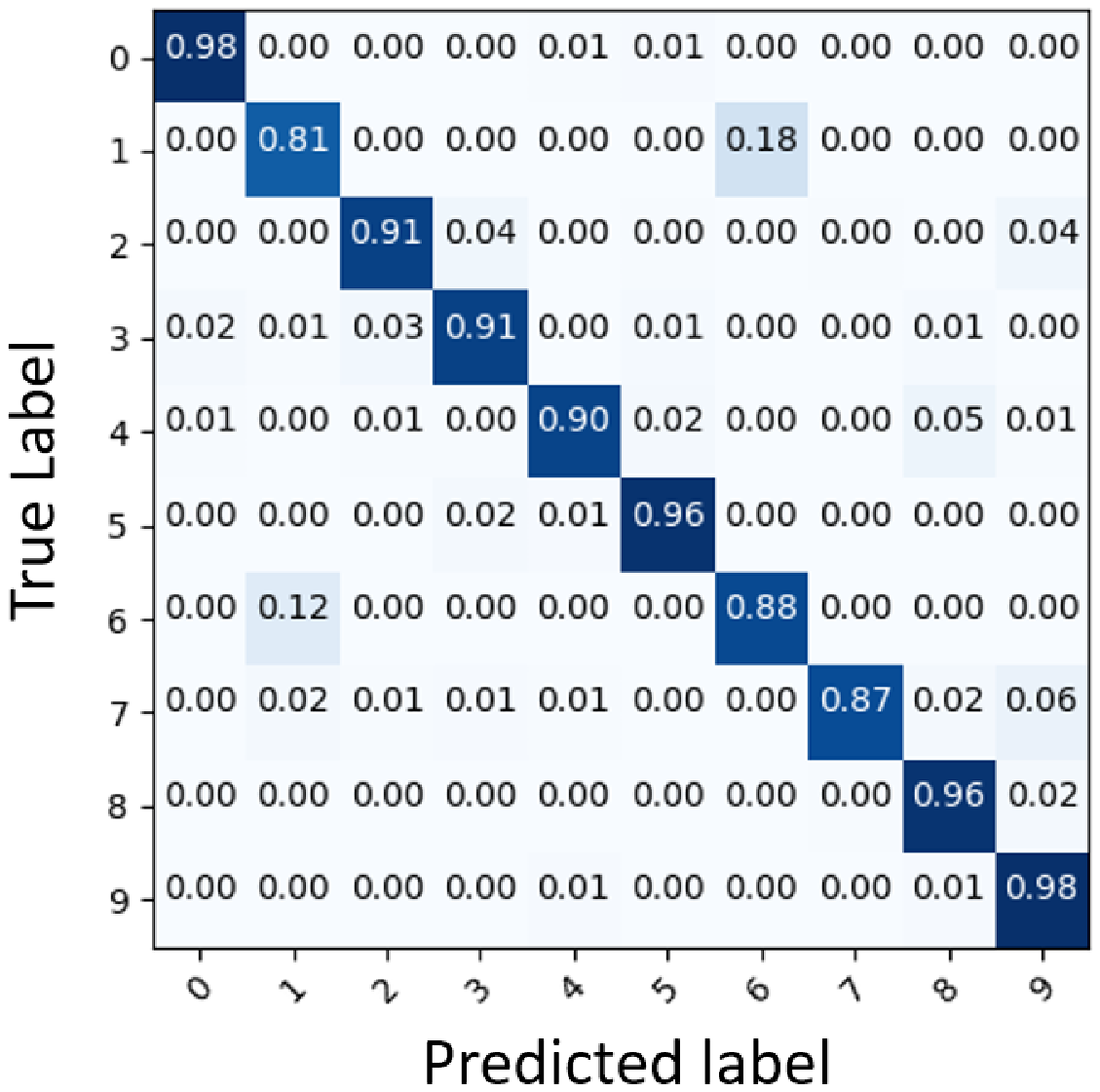}}
  \quad
  \quad
  \subfloat[per image \label{fig:cm_2_img}]{\includegraphics[width=0.44\linewidth]{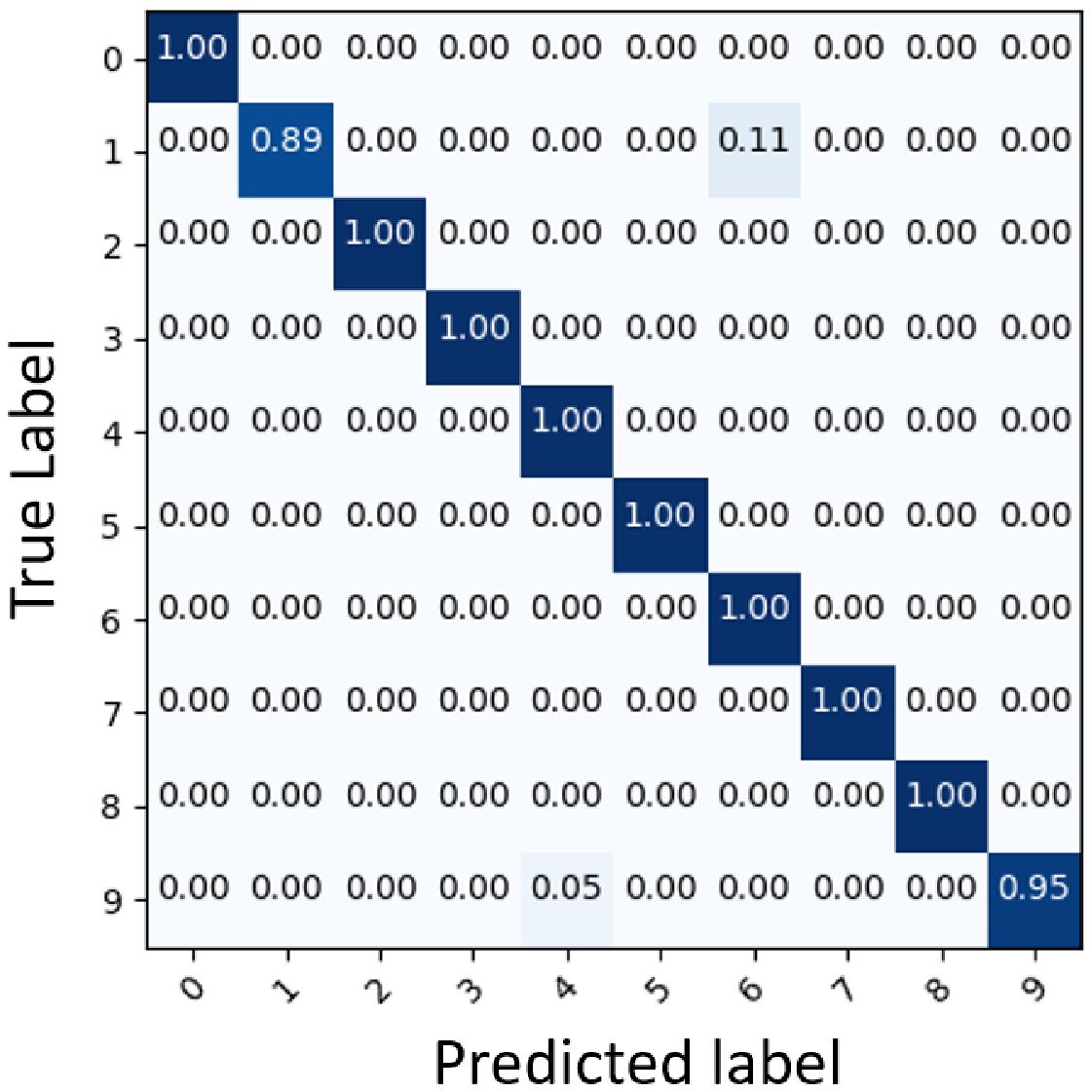}}

  \caption{Confusion matrices for the 10-scanner dataset. 
  Each cell indicates the classification accuracy of predicated labels based on true labels. 
  }
  \label{fig:cm_2}
\end{figure}

\subsection{Testing}
\label{ssec:testing}
The same pre-processing procedure as described in the training section will be used in the testing stage. 
A test image will first be split into sub-images, and then subsequently extracted into patches of size $64 \times 64$ pixels.
The extracted patches will be used as inputs for the proposed neural network.

As shown in Figure \ref{fig:system}, our proposed system will evaluate two tasks on scanned images: scanner model classification and reliability map generation. 
In Task 1 (scanner model classification), we assign the predicted scanner labels to both patches $I_{p}$ and original images $I$. 
The predicted scanner label for the sub-image $I_{s}$ is the same as the predicted label of its corresponding patch $I_{p}$.
The classification decision for the original image $I$ is obtained by majority voting over the decisions corresponding to its individual sub-images $I_{s}$.
In Task 2, a reliability map~\cite{zhou2015cnnlocalization} is generated based on the majority vote result from Task 1.
The pixel values in the reliability map indicate the probability of the corresponding pixel in the original image being correctly classified.
The probability of pixel $x$ belonging to scanner $s$ is the average value of the corresponding probabilities for the sub-images that contain pixel $x$:
\begin{equation}
\label{eq:mv}
  \mathcal{P}_s(x) = \frac{1}{n} \sum^{n}_{i=1} \mathcal{P}_s(Sub_i)
\end{equation}
where $Sub_i$ indicates the sub-image including pixel $x$, $n$ indicates the total number of these sub-images, and $\mathcal{P}_s(\cdot)$ indicates the probability of $\cdot$ belongs to scanner $s$.

\section{The Experiments}
\label{sec:exp}
In this section, we describe the dataset we use and the experiments conducted by using the proposed system in Figure \ref{fig:system}.

\subsection{Dataset}
\label{ssec:dataset}
We use the Dartmouth Scanner Dataset for our experiments.\footnote{We like to thank Professor Hany Farid for making this dataset available.}
This dataset contains a total of \SI{3874 } scanned images in JPEG format from \SI{169 } different scanner models. 
The size of the original scanned images varies from $500 \times 500 $ pixels to $5,000 \times 5,000$ pixels with various scan resolutions (dpi - dots per inch).  
For each scanner model, we randomly partition its images into three subsets with the ratio of 6:1:3 for training, validation, and testing, respectively.
We first construct a small sub-dataset with 10 randomly selected scanners, known as the ``10-scanner dataset'', to evaluate the performance of the proposed system.
We then use the entire dataset to check whether the system is able to scale to a larger dataset.

\begin{table}[!hbt]
  \centering
  \begin{tabular}{ |c|c|c|c| }
    \hline
    Network &  & 10 scanners & 169 scanners \\ \hline
    \multirow{2}{*}{Ours}      & per image & \red{$96.83 \%$} & $92.97 \%$\\
                      & per patch & \red{$93.72 \%$} & \red{$89.69 \%$}\\ \hline
    \multirow{2}{*}{Xception~\cite{xception}}      & per image & $95.24 \%$ & \red{$93.24 \%$}\\
                      & per patch & $92.11 \%$ & $88.85 \%$\\ \hline
    \multirow{2}{*}{Inception3~\cite{szegedy_2016}}      & per image & $94.44 \%$ & $90.37 \%$\\
                      & per patch & $91.69 \%$ & $88.62 \%$\\ \hline
    \multirow{2}{*}{Resnet34~\cite{resnet}}      & per image & $96.03 \%$ & $91.67 \%$\\
                      & per patch & $91.72 \%$ & $88.73 \%$\\ \hline
  \end{tabular}
  \caption{The scanner model classification accuracy: ``per patch'' rows indicate the classification accuracy on patches $I_{p}$; ``per image'' rows indicate the classification accuracy for full size images $I$.
  }
  \label{tab:comp_2}
\end{table}

We also constructed several forged images using a copy-move attack for evaluating our reliability maps. The copied areas are from the same image (``self copy'') or from a different image in the dataset (``multi-source copy').

\subsection{Experimental Results}
\label{ssec:setup}

{\bf Task 1 --- Scanner model classification.} 
Our neural network (Figure \ref{fig:patch_incep_res}) is implemented in Pytorch using stochastic gradient descent (SGD) with learning rate 0.01, momentum 0.5 and weight decay 0.0001.
We compare our method with some other CNN architectures, such as InceptionV3~\cite{szegedy_2016}, Resnet34~\cite{resnet} and Xception~\cite{xception}.


Figure \ref{fig:cm_2} reports our results in terms of the confusion matrices for the 10-scanner dataset. 
The overall classification accuracy is $93.72 \%$ per patch (i.e.\ without majority vote) and $96.83 \%$ per image (i.e.\ with majority vote). 
The high accuracies on patch-level and image-level classification tasks indicate our model is very effective on the 10-scanner dataset.
The results for both the 10-scanner dataset and the entire dataset are reported in Table \ref{tab:comp_2}.
On the 10-scanner dataset, our method achieves the highest classification accuracy on both patches and the images.
On the entire dataset, our method achieves the highest patch-level accuracy.
Our image-level classification accuracy is very close to the highest, the one which achieved by Xception. 
It must be noted that our model has fewer parameters and is shallower compared to the other CNN architectures.

\begin{figure}[!htb]
  \centering 
  \begin{minipage}[b]{1.0\linewidth}
    \centering
    \includegraphics[width=0.45\linewidth,valign=c]{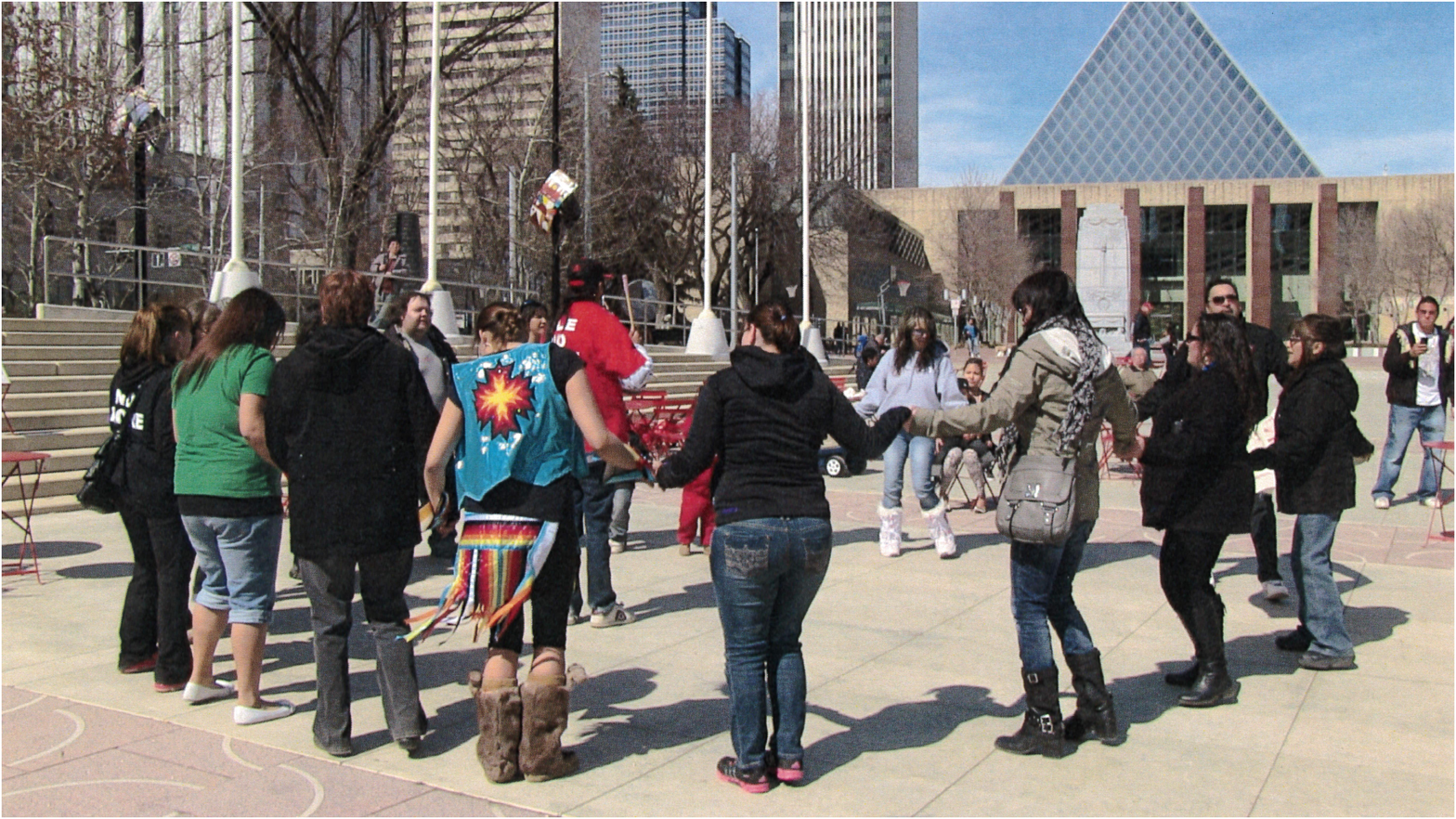} 
    \includegraphics[width=0.52\linewidth,valign=c]{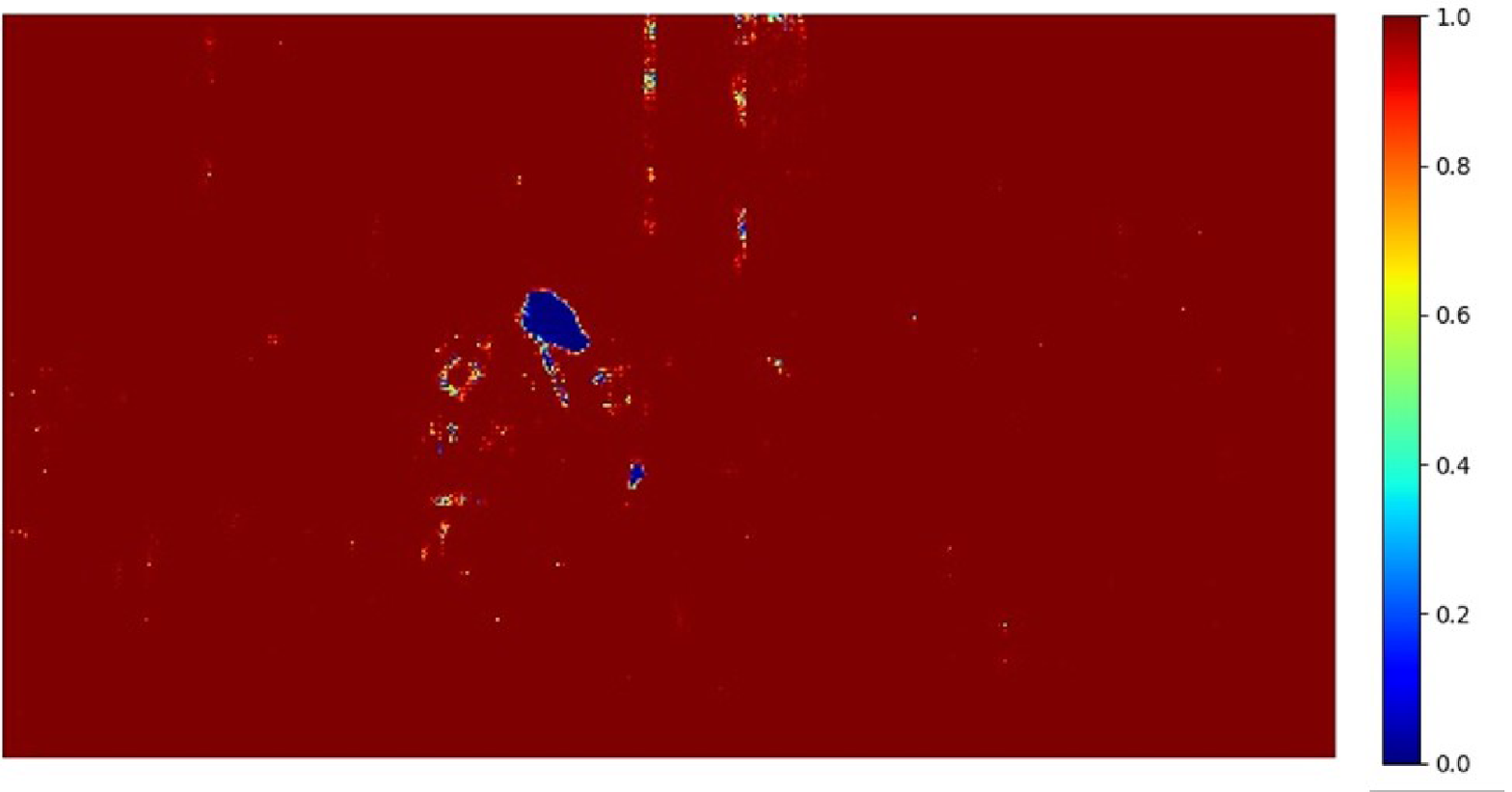} 
  \end{minipage}
  \caption{An original scanned image used for forged image creation and its corresponding reliability map with sub-images size set to be $64 \times 64$ pixels and stride set to be 4 pixels.}
  \label{fig:fig8}
\end{figure}

\begin{figure*}[!htb]
  \centering
  \begin{minipage}[b]{0.16\linewidth}
    \centering
    \centerline{\includegraphics[width=\linewidth]{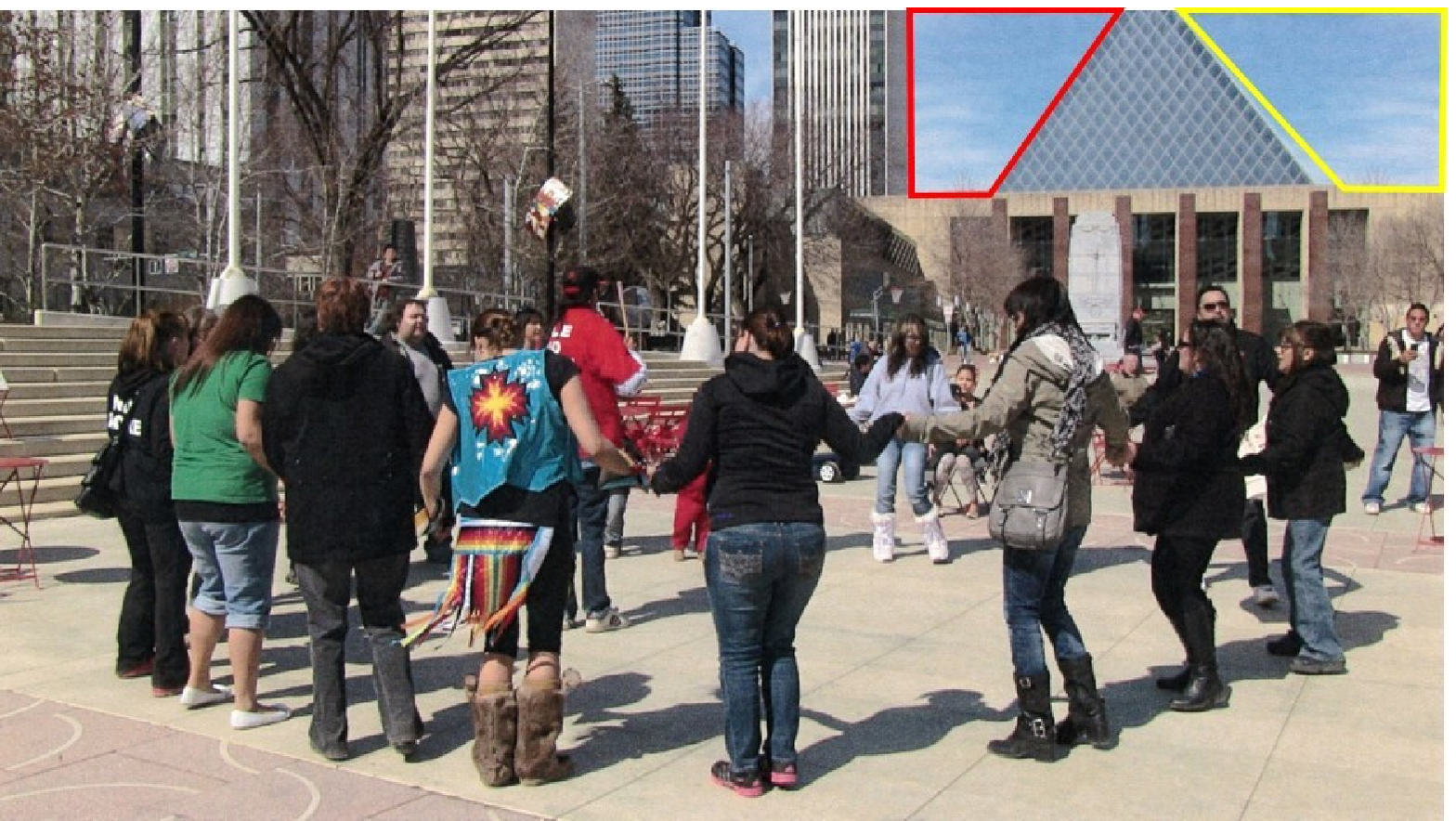}}
    \vspace{2pt}
    \centerline{\includegraphics[width=\linewidth]{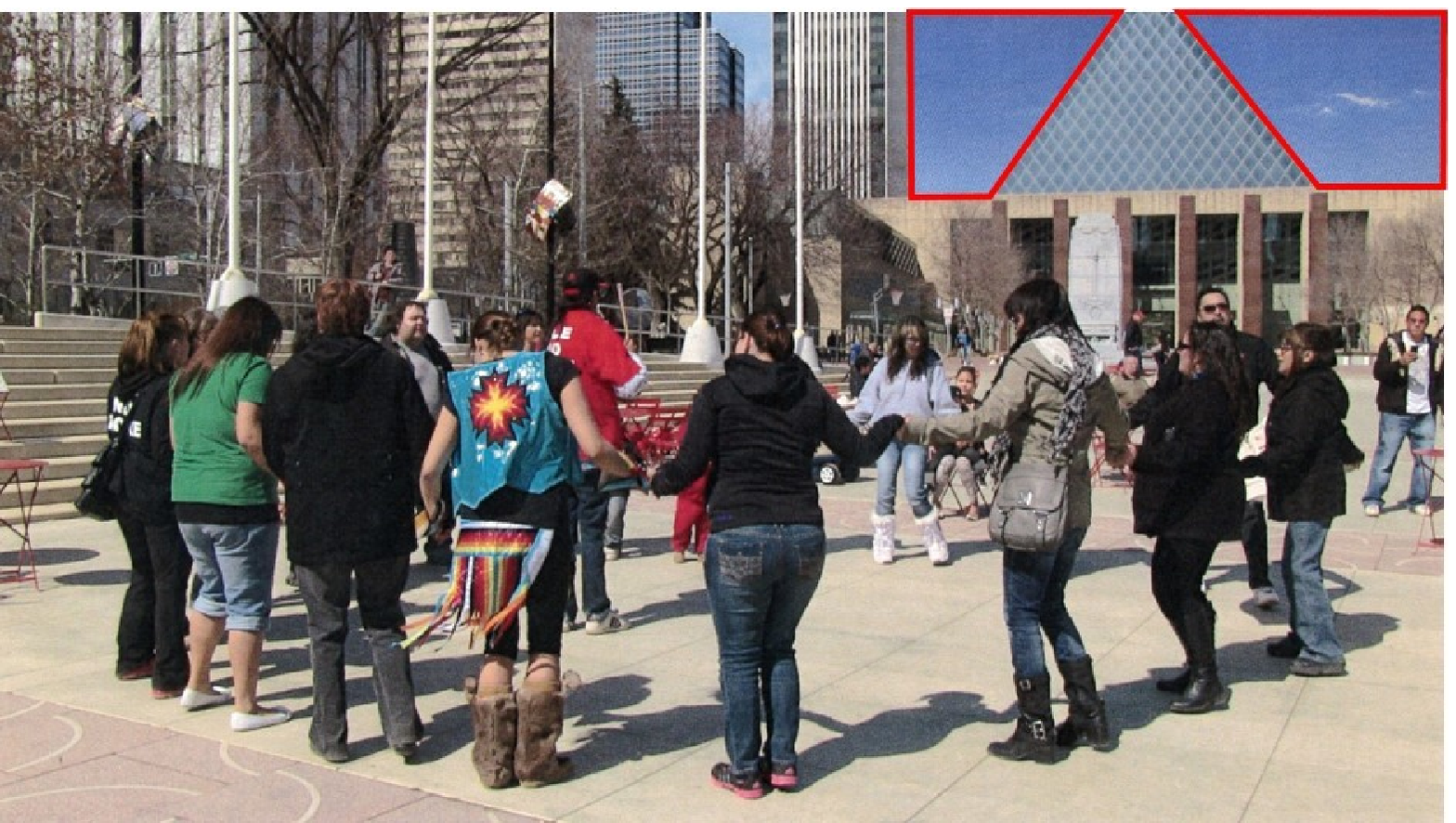}}
    \centerline{Forged Image}\medskip
  \end{minipage}
  \hfill
  \begin{minipage}[b]{0.16\linewidth}
    \centering
    \centerline{\includegraphics[width=\linewidth]{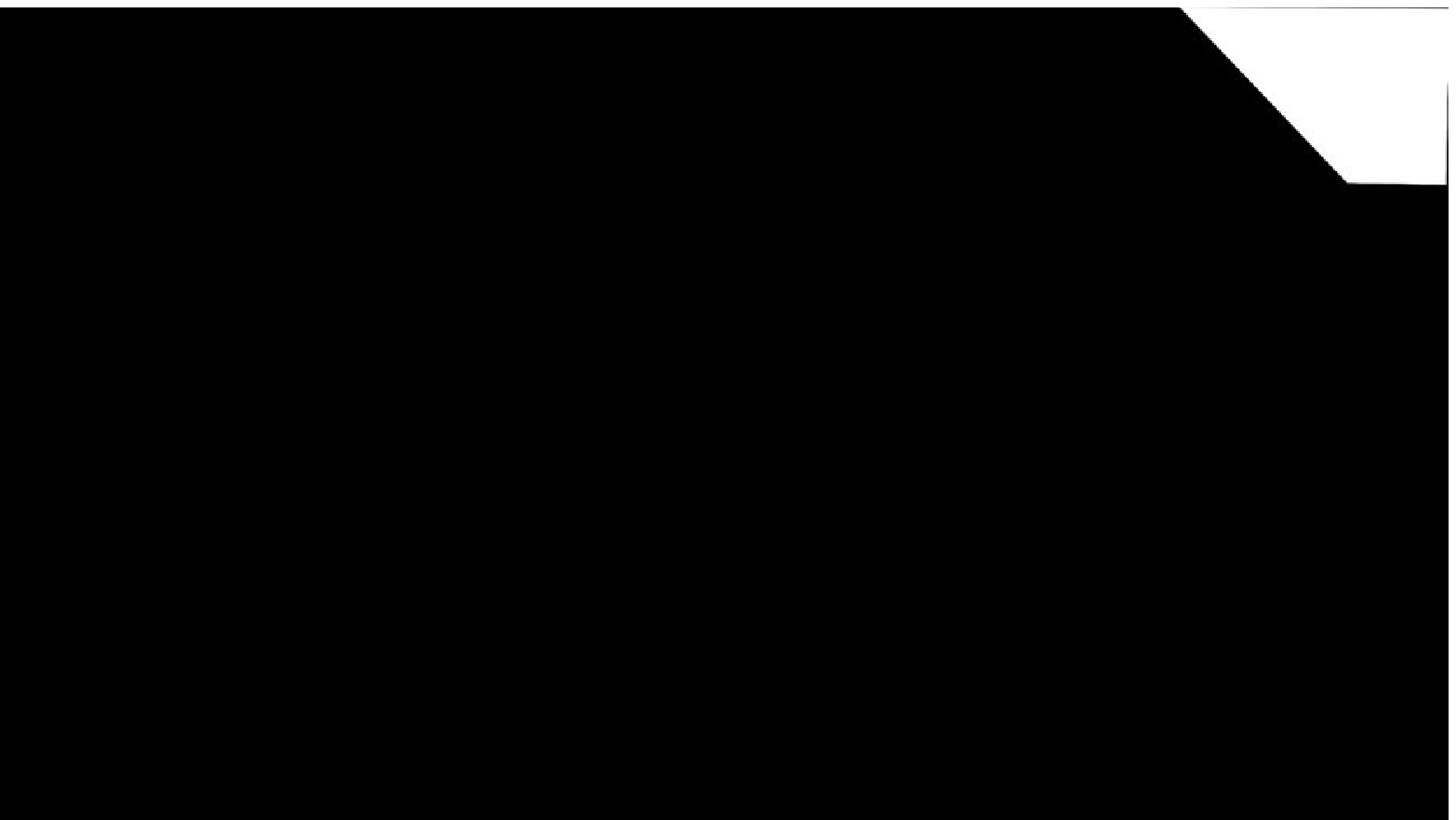}}
    \vspace{2pt}
    \centerline{\includegraphics[width=\linewidth]{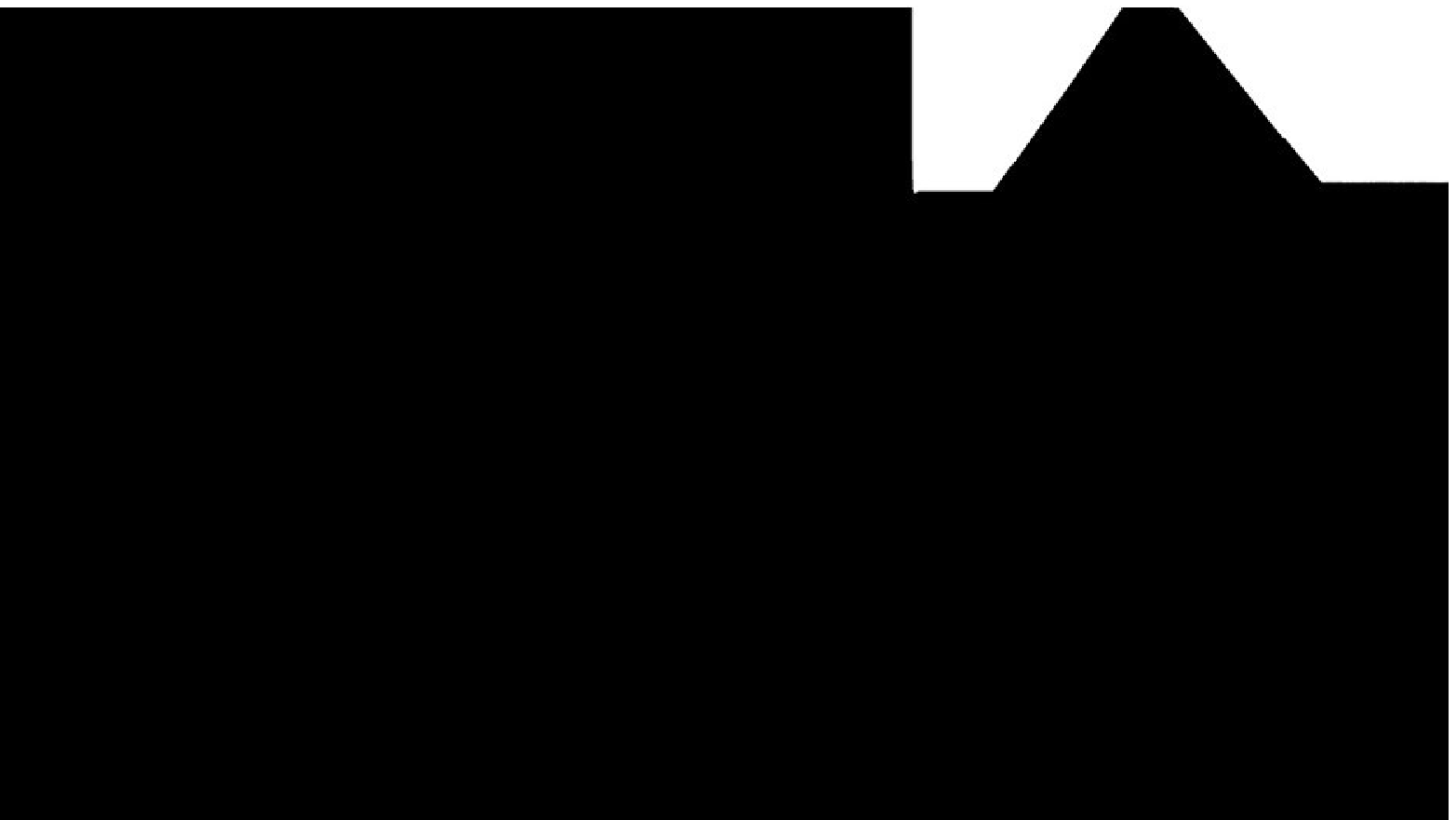}}
    \centerline{Ground Truth}\medskip
  \end{minipage}
  \hfill
  \begin{minipage}[b]{0.16\linewidth}
    \centering
    \centerline{\includegraphics[width=\linewidth]{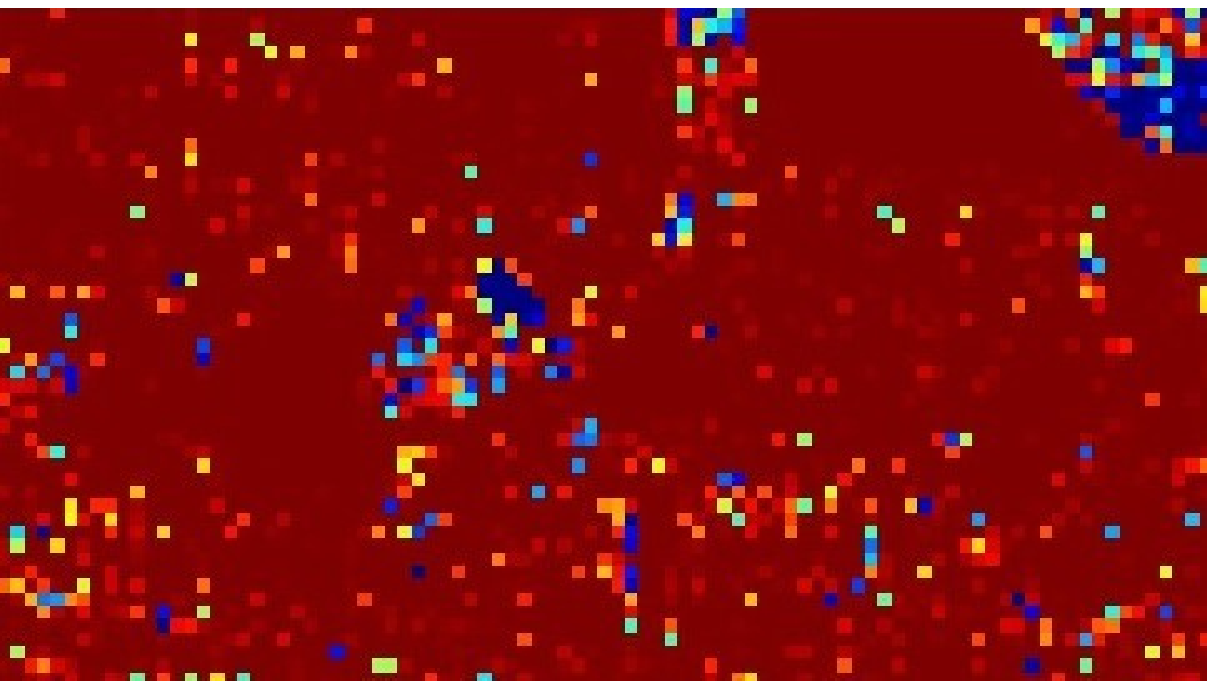}}
    \vspace{2pt}
    \centerline{\includegraphics[width=\linewidth]{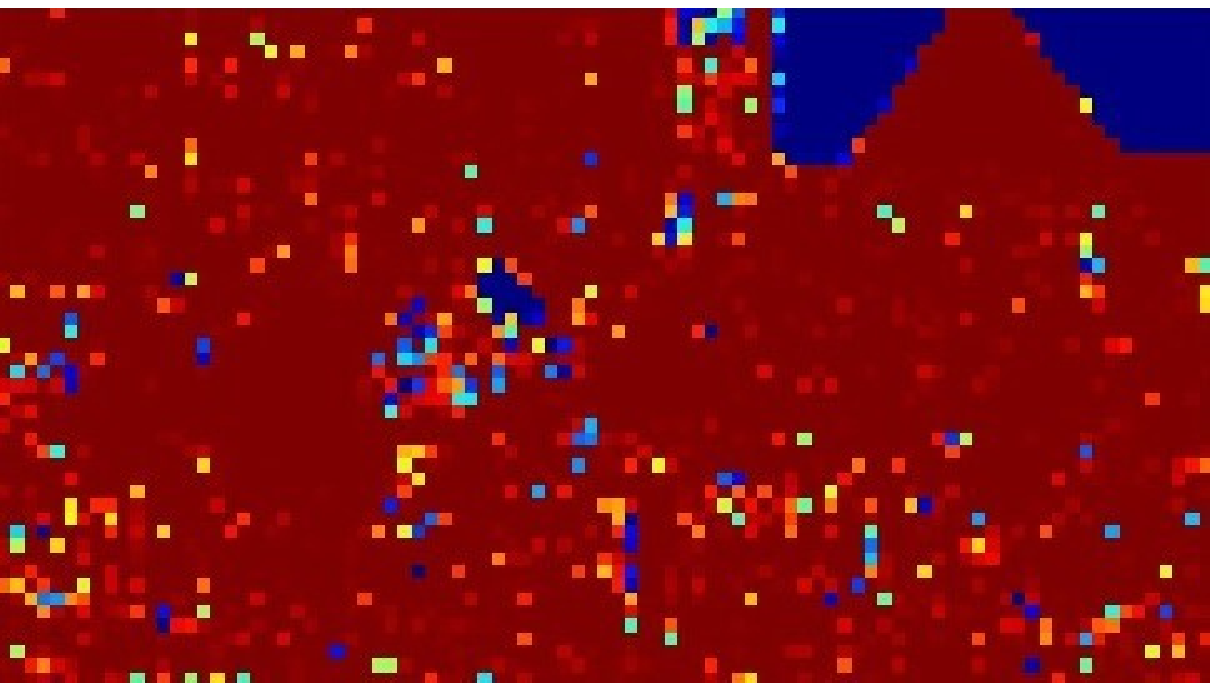}}
    \centerline{Stride 64}\medskip
  \end{minipage}
  \hfill
  \begin{minipage}[b]{0.16\linewidth}
    \centering
    \centerline{\includegraphics[width=\linewidth]{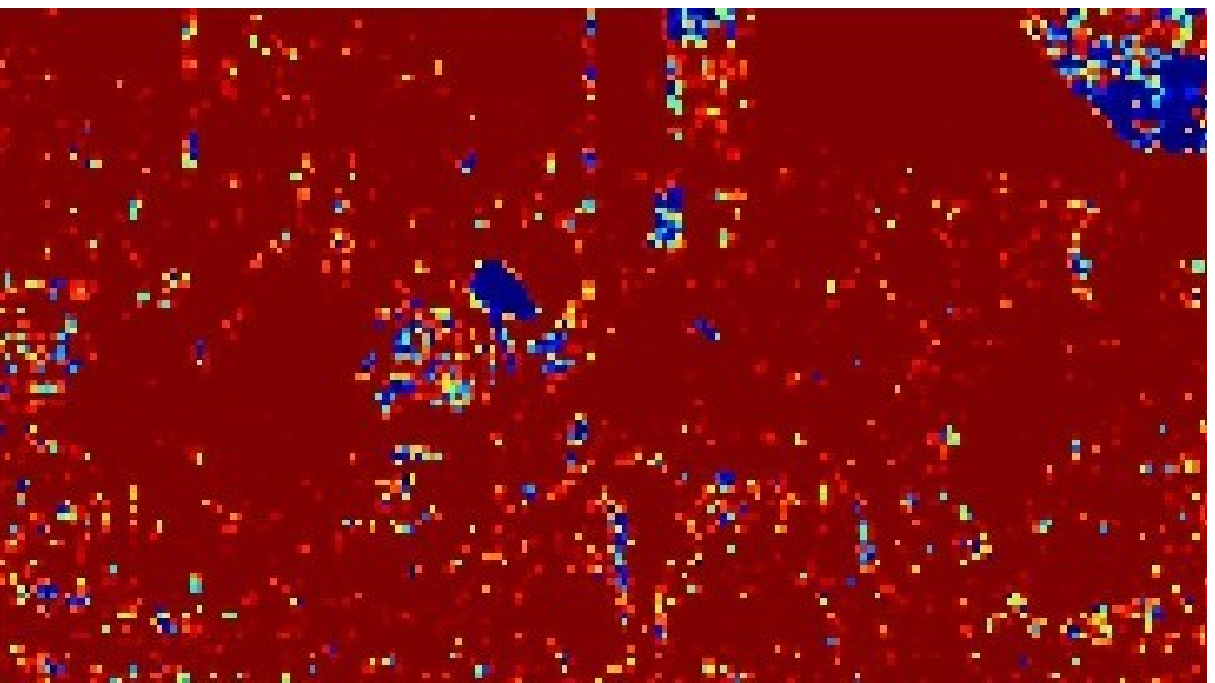}}
    \vspace{2pt}
    \centerline{\includegraphics[width=\linewidth]{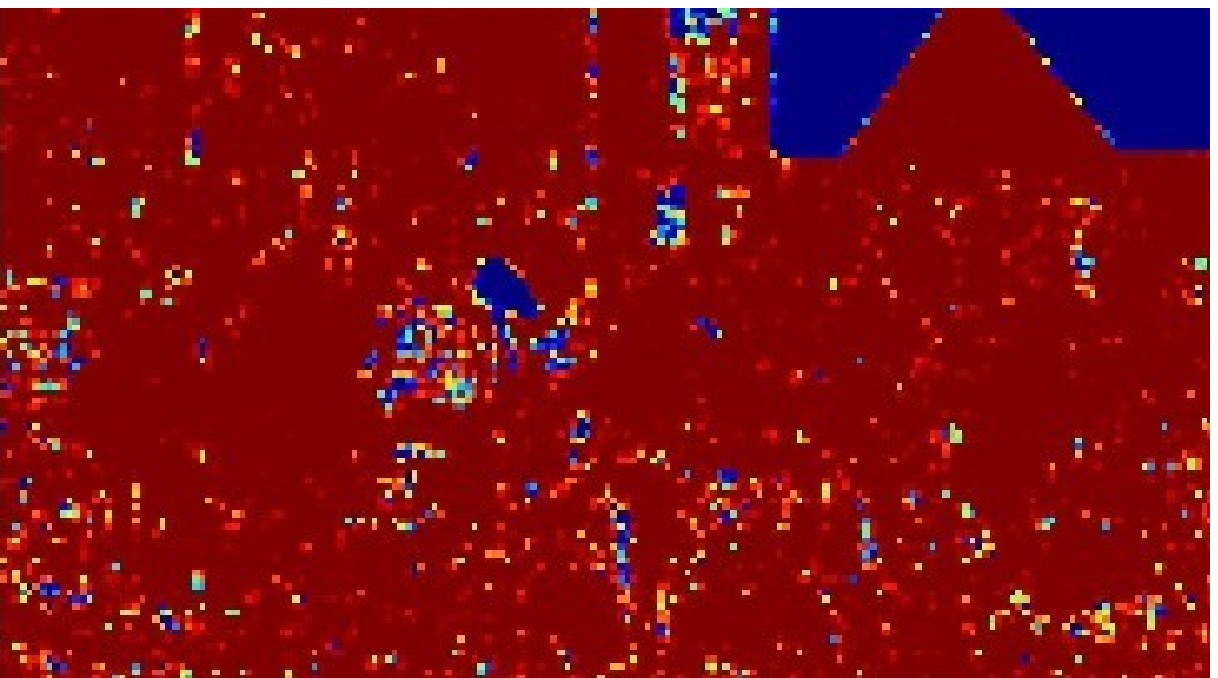}}
    \centerline{Stride 32}\medskip
  \end{minipage}
  \hfill
  \begin{minipage}[b]{0.16\linewidth}
    \centering
    \centerline{\includegraphics[width=\linewidth]{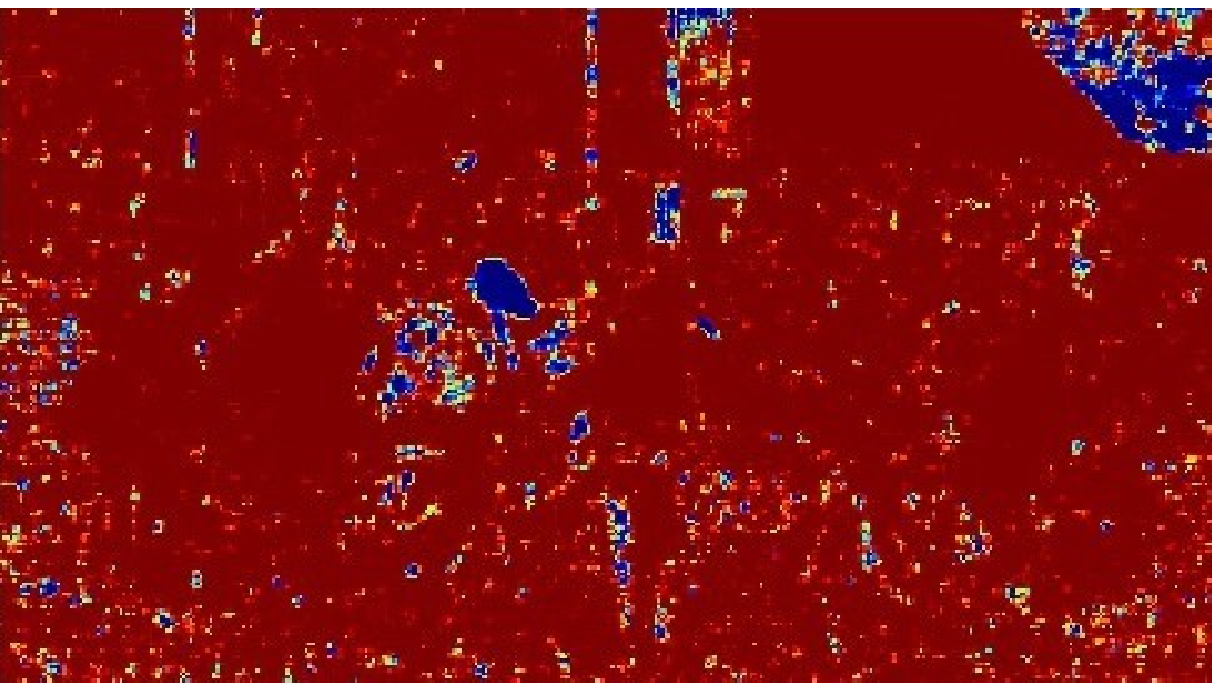}}
    \vspace{2pt}
    \centerline{\includegraphics[width=\linewidth]{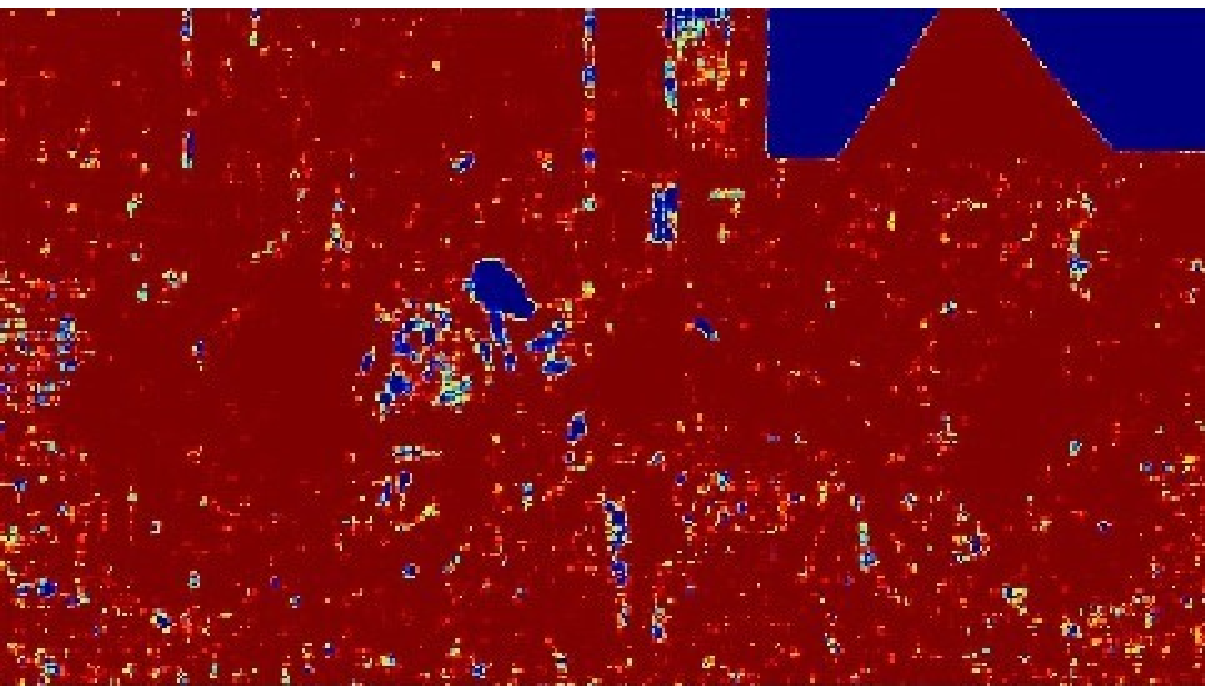}}
    \centerline{Stride 16}\medskip
  \end{minipage}
  \hfill
  \begin{minipage}[b]{0.16\linewidth}
    \centering
    \centerline{\includegraphics[width=\linewidth]{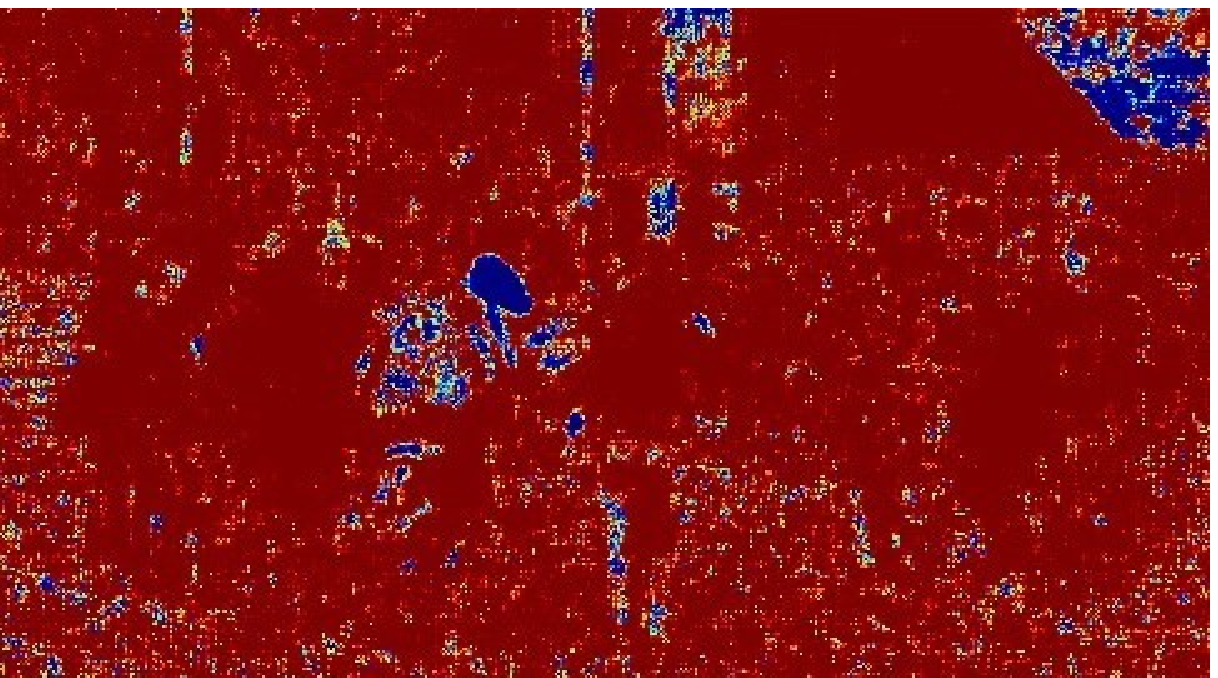}}
    \vspace{2pt}
    \centerline{\includegraphics[width=\linewidth]{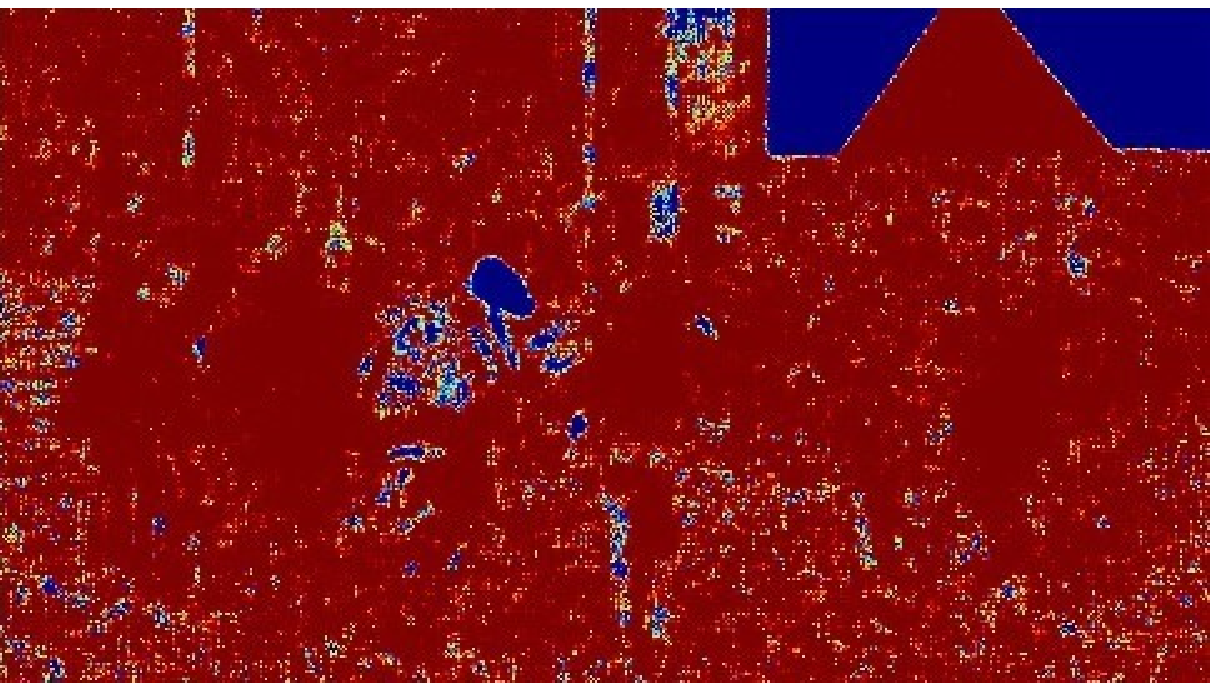}}
    \centerline{Stride 4}\medskip
  \end{minipage}
  \caption[]{The forged scanned images and corresponding reliability maps with different strides. 
  \begin{enumerate*}[label=(\roman*)]
  \item For the self copy forged image (top), the yellow box region is duplicated from the red box region with horizontal flipping, stretching and compressing operation;
  \item For the multi-source forged image (bottom), the red box region is duplicated from another image scanned by different scanner.
  \end{enumerate*}
  As the reliability maps shows, the smaller the stride, the reliability map can achieve better localization.}
  \label{fig:forged}
\end{figure*}

{\bf Task 2 --- Reliability Maps.}
Since our system is aimed at extracting intrinsic features of scanner models, it should also be able to identify manipulated region irrespective of image content.
In this task, we investigate to generate a reliability map (\ie a heat map) that can indicate suspicious forged areas in the images.
The reliability map is generated based on the predicted label obtained by majority vote, as explained in equation \ref{eq:mv}.

Figure \ref{fig:fig8} shows an example of the reliability map.
In the reliability map, the color of the pixel represents the probability that it is generated by the predicted scanner model. 
Color ``dark red'' indicates a probability value equal to $1.0$, and color ``dark blue'' indicates a probability value equal to $0.0$.
Then we use the original image from Figure \ref{fig:fig8} to generate manipulated images in Photoshop.
The forged images are shown in the first column in Figure \ref{fig:forged}.
The top one is generated by self-image copy-move with translation operations.
The bottom one is generated by copy-pasting regions in an other image source from different scanner model. 
The reliability maps generated with different stride size for these two forged images are shown in Figure \ref{fig:forged}. 
These results indicate the effectiveness of using our reliability map to indicate the suspicious forgery.

\section{Conclusion}
\label{sec:cons}
In this paper we investigate the use of deep-learning methods to address  scanner model classification and localization. 
Compared with classical methods, our proposed system can: 
\begin{enumerate*}[label={\alph*)}]
  \item learn intrinsic scanner features automatically;
  \item have no restrictions on data collection; 
  \item associate small image patches ($64 \times 64$ pixels) to scanner models with high accuracy; and
  \item detect image forgery and localization on small image size.
\end{enumerate*}
Our experimental results shown in Table \ref{tab:comp_2} indicate that the proposed system can automatically learn the inherit features to differentiate scanner models and is robust to JPEG compression.
The results in Figure \ref{fig:forged} show the ability of the proposed system to identify suspected forged regions in scanned images. 
These experimental results indicate that our reliability map provides a way to detect forgeries in scanned images.

Further work will be devoted to:
\begin{enumerate*}[label={\alph*)}]
  \item improve the neural network architecture in the proposed system, and
  \item evaluate the performance of the proposed system on scanned documents.
\end{enumerate*}

\section{Acknowledgments}
This material is based on research sponsored by the Defense Advanced Research Projects Agency (DARPA) and the Air Force Research Laboratory (AFRL) under agreement number FA8750-16-2-0173.
The U.S. Government is authorized to reproduce and distribute reprints for Governmental purposes notwithstanding any copyright notation thereon.
The views and conclusions contained herein are those of the authors and should not be interpreted as necessarily representing the official policies or endorsements, either expressed or implied, of DARPA, AFRL or the U.S. Government. Address all comments to Edward J. Delp, ace@ecn.purdue.edu.

\bibliographystyle{IEEEtran}
\bibliography{reference}

\end{document}